\documentclass[letterpaper, 10 pt, conference]{ieeeconf}  

\IEEEoverridecommandlockouts                              

\usepackage{graphicx}
\usepackage{amsmath}
\usepackage{amssymb}
\usepackage{xcolor}
\usepackage{booktabs}
\usepackage{subcaption}
\usepackage{hyperref}
\usepackage{url}

\title{\LARGE \bf Baby Sophia: A Developmental Approach to Self-Exploration \\ through Self-Touch and Hand Regard}

\author{Stelios Zarifis$^{1,2,*}$, Ioannis Chalkiadakis$^{3}$, Artemis Chardouveli$^{3}$, Vasiliki Moutzouri$^{3}$, Aggelos Sotirchos$^{3}$, \\ Katerina Papadimitriou$^{1,2}$, Panagiotis Filntisis$^{1,2}$, Niki Efthymiou$^{1,2}$, Petros Maragos$^{1,2}$ and Katerina Pastra$^{3}$
	\thanks{$^{1}$Robotics Institute, Athena Research Center, Athens, Greece}
	\thanks{$^{2}$HERON -- Hellenic Robotics Center of Excellence, Athens, Greece}
	\thanks{$^{3}$Institute for Language and Speech Processing, Athena Research Center, Athens, Greece}
	\thanks{$^{*}$Corresponding author: \tt\footnotesize s.zarifis@athenarc.gr}
	\thanks{The code and experiments for this work are available at \url{https://github.com/SteliosZarifis/BabyBenchAthena2025}}
}

\begin{document}
	\maketitle
	\thispagestyle{empty}
	\pagestyle{empty}
	
	\begin{abstract}
		Inspired by infant development, we propose a Reinforcement Learning (RL) framework for autonomous self-exploration in a robotic agent, Baby Sophia, using the BabyBench simulation environment~\cite{lopez2025babybench}. The agent learns self-touch and hand regard behaviors through intrinsic rewards that mimic an infant's curiosity-driven exploration of its own body. For self-touch, high-dimensional tactile inputs are transformed into compact, meaningful representations, enabling efficient learning. The agent then discovers new tactile contacts through intrinsic rewards and curriculum learning that encourage broad body coverage, balance, and generalization. For hand regard, visual features of the hands, such as skin-color and shape, are learned through motor babbling. Then, intrinsic rewards encourage the agent to perform novel hand motions, and follow its hands with its gaze. A curriculum learning setup from single-hand to dual-hand training allows the agent to reach complex visual-motor coordination. The results of this work demonstrate that purely curiosity-based signals, with no external supervision, can drive coordinated multimodal learning, imitating an infant's progression from random motor babbling to purposeful behaviors.
	\end{abstract}
	
	\vspace{.5em}
	\section{Introduction}
	
	In the first months of life, human infants engage in motor babbling—producing movements that appear random and uncoordinated. Despite their seemingly chaotic nature, these actions generate rich sensorimotor data that becomes the foundation for later organized motor control and body awareness~\cite{rochat1998self, mannella2018know}. Two particularly important early behaviors are self-touch and hand regard, which play crucial roles in developing an internal body schema and coordinating visual-motor systems.
	
	\vspace{.5em}
	\subsection{Self-Touch in Infant Development}
	
	Self-touch is one of the earliest forms of self-exploration in infants. Each episode of self-touch provides dual sensory feedback: proprioceptive signals from the moving limb indicating its pose in space, and tactile signals from the points of contact. With this feedback mechanism the agent gradually forms an internal body schema—a representation of its own body in space~\cite{thomas2015independent, dimercurio2018naturalistic, khoury2022self}.
	
	Empirical studies have shown that self-touch in infants follows a systematic developmental pattern. Thomas et al.~\cite{thomas2015independent} documented that self-touch progresses in a rostral-to-caudal (head-to-tail) order. Infants begin by touching their face and head regions, then progress to their torso, and later extend their exploration to their hips, legs, and eventually feet. This pattern is not random but reflects the developmental maturation of motor control systems, which proceeds from the head downward.
	
	An important milestone in self-touch development is midline crossing—the ability to reach across the body's midline with one hand to touch body parts on the opposite side~\cite{dimercurio2018naturalistic}. This behavior emerges around 3-4 months of age and signals improved bilateral coordination and body awareness. Additionally, as infants gain better head and neck control, they bring distal body parts like their feet close to their face, enabling early exploration of these harder-to-reach regions~\cite{khoury2022self}. Inspired by this behavior, we incorporate a similar milestone reward to motivate the infant-robot to reach new body parts.
	
	Initially, self-touch movements are largely random, driven by spontaneous motor activity. However, over developmental time, these actions become increasingly goal-directed and purposeful. This transformation is believed to be guided by intrinsic motivation—infants' inherent drive to explore novelty and seek learning progress~\cite{mannella2018know, gama2021goal}. The infant brain effectively rewards itself for discovering new sensations and achieving new motor competencies, without requiring external reinforcement.
	
	\vspace{.5em}
	\subsection{Hand Regard in Infant Development}
	
	Hand regard is another fundamental behavior in early infant development. It typically appears around the second month of life, reaches peak frequency in the third month, and then gradually declines by the fifth month~\cite{white1964observations, rochat2004infant}. During hand regard episodes, infants bring their hands into their field of view and fixate their gaze upon them, often slowly rotating and moving their hands while watching intently. This behavior combines visual feedback (seeing the hand move) with proprioceptive feedback (feeling the hand move), creating a powerful multimodal learning experience~\cite{corbetta2018embodied}.
	
	Hand regard serves as a developmental bridge between early self-exploration and purposeful interaction with the environment. By the fourth month, as hand regard begins to decline, it is replaced by visually guided reaching—infants start to quickly move their hands toward objects they see, integrating visual information with motor planning~\cite{corbetta2021perception}. This transition marks a critical shift from self-focused exploration to world-focused interaction.
	
	The function of hand regard extends beyond simple visual exploration. It strengthens multisensory integration, linking visual appearance with proprioceptive sensation and motor commands. It also supports the emergence of self-agency—the understanding that one's own actions produce observable effects in the world. Infants learn the relation between motor commands (moving the arm), proprioceptive sensations (feeling the arm move), and visual consequences (seeing the hand move)~\cite{rochat2004infant}.
	
	\vspace{.5em}
	\subsection{Contributions}
	
	Inspired by these developmental principles, we present \emph{Baby Sophia}, a reinforcement learning agent in the BabyBench simulator~\cite{lopez2025babybench} that autonomously learns both behaviors using only intrinsic rewards (no external or task-specific supervision). Our main contributions are:
	
	For self-touch:
	\begin{itemize}
		\item The use of a semantic body map compressing the prohibitively high-dimensional tactile vector into meaningful, denser representations.
		\item A multi-component intrinsic reward driving broad, symmetric exploration with infant-like progression.
		\item A two-stage curriculum improving robustness, and accelerating learning.
	\end{itemize}
	
	For hand regard:
	\begin{itemize}
		\item The use of motor babbling to autonomously understand hand appearance via the correlation of proprioception and visual signals.
		\item Curiosity-based visual rewards that sustain hand tracking.
	\end{itemize}
	
	We show that intrinsic motivation and curriculum learning (widely studied and well-understood mechanisms of infant development) enable artificial agents to acquire complex sensorimotor skills from raw, high-dimensional inputs. This work demonstrates that curiosity alone can successfully drive embodied learning.
	
	\vspace{.5em}
	\section{Methodology}
	\label{sec:methodology}
	We use the BabyBench framework~\cite{lopez2025babybench}, which provides a simulation for a robotic infant model (MIMo) with realistic physics, high-dimensional sensory observations, and continuous motor control. We employ Proximal Policy Optimization (PPO)~\cite{schulman2017proximalpolicyoptimizationalgorithms}, a reinforcement learning algorithm, to enable the agent to learn motor policies $\pi(a_t|s_t)$ to control its body parts and achieve the desired behaviors. The learning process is driven entirely by intrinsic reward signals $R_t$, designed to encourage exploration, inspired by developmental psychology, where the agent initially engages in random movements (motor babbling) and gradually acquires finer, coordinated skills through progressive stages~\cite{rochat1998self, bengio2009curriculum}, without external task supervision.
	
	\vspace{.5em}
	\subsection{Model I: Self-Touch via Tactile Novelty}
	\label{subsec:self_touch}
	We develop a method for the agent to learn the self-touch behavior by constructing a compact body map from the high-dimensional tactile signals. Then, driven by intrinsic rewards, the agent discovers its own body by being curious about sensor areas it has not discovered before.
	
	\subsubsection{Handling High-Dimensions}
	The agent receives raw tactile observations $o^{\text{Touch}}_t \in \mathbb{R}^{17,175}$, representing sensor activations across the body. This high dimensionality poses challenges:
	\begin{itemize}
		\item Sample Complexity: The large state-action space makes exploration intractable.
		\item Noise: Individual sensors produce highly variable signals—fluctuating rapidly in a wide range of values due to minor contacts or even simulation noise—making it extremely difficult for the policy to learn efficient mappings from tactile input to motor control.
	\end{itemize}
	
	\subsubsection{Body Map Construction}
	To address this challenge, we compress the tactile space into a body map, grouping the $17,175$ sensors into $G=68$ anatomical regions ($34$ body parts per side: face ($10$), torso ($14$), arms ($18$), legs ($18$), feet ($8$)). For each group $S_j$, we compute the mean activation signal:
	\begin{equation*}
		g_{t,j} = \frac{1}{|S_j|}\sum_{i \in S_j} |o^{\text{Touch}}_{t,i}|, \quad g_t \in \mathbb{R}^{68}
	\end{equation*}
	In this way, we achieve both a smaller observation dimension and a more semantically meaningful representation. By aggregating sensors into body regions, we provide the agent with one signal per functional area—mimicking human tactile perception, where we experience touch in terms of body parts (e.g., "arm", "forearm", "hand") rather than individual sensor points. The tactile vector $g_t$ is processed by a two-layer MLP ($68 \rightarrow 128 \rightarrow 256$) to produce a representation $h^{\text{touch}}_t$. Proprioceptive data $o^{\text{Proprio}}_t$ is similarly processed to $256$ dimensions, and both are fused via a linear layer with ReLU into $h_t \in \mathbb{R}^{512}$, encoding touch and body configuration.
	
	\subsubsection{Intrinsic Reward Design}
	The total intrinsic reward $R_t$ combines four components for systematic, balanced, and sustained self-exploration:
	\begin{equation*}
		R_t = R_{\text{touch}} + R_{\text{geom}} + R_{\text{milestones}} + R_{\text{balance}}
	\end{equation*}
	\begin{itemize}
		\item Touch Novelty ($R_{\text{touch}}$): Encourages discovering and briefly exploring each body part. A body part $j$ is marked as touched if $g_{t,j} > 0.1$. Rewards depend on global and episode-level novelty:
		\begin{equation*}
			R_{\text{touch}}^j(t) =
			\begin{cases}
				5.0, \quad \text{first-ever touch} \\
				1.0, \quad \text{first in current episode} \\
				0.5 \cdot f_{\text{decay}}(c_j^{\text{episode}}), \text{otherwise}
			\end{cases}
		\end{equation*}
		with $f_{\text{decay}}(c) = \frac{2}{1 + \exp(0.5(c - 10))} - 1$. This is the \emph{anti-boredom} mechanism which initially rewards repeated contact to support local exploration, then gradually reduces and eventually crosses zero and penalizes fixation ($f_{\text{decay}} \to -1$), pushing the agent to move on when it is bored.
		\item Contact Novelty ($R_{\text{geom}}$): Rewards the \emph{event} of a specific hand $h$ making first contact with body part $j$:
		\begin{equation*}
			R_{\text{geom}}^{(h,j)}(t) =
			\begin{cases}
				10.0 & \text{first global contact} \\
				2.0 & \text{first in episode}
			\end{cases}
		\end{equation*}
		This sparse, high-value signal reinforces trajectories that successfully reach a new area.
		\item Diversity Milestones ($R_{\text{milestones}}$): One-time bonuses of $5.0$ when the number of unique body parts touched in an episode crosses thresholds $\{5, 10, 15, 20, 25, 30\}$. Encourages broad coverage.
		\item Balance Reward ($R_{\text{balance}}$): A $10.0$ bonus at episode end if $|N_L - N_R| \leq 3$, where $N_L$ and $N_R$ are the number of distinct parts touched by left and right hands. Promotes ambidextrous exploration.
	\end{itemize}
	Combined, these rewards create a natural progression, as a human learns increasingly difficult skills, driven purely by intrinsic motivation.
	
	\subsubsection{Policy Network Architecture}
	The fused representation $h_t \in \mathbb{R}^{512}$ is processed by two-layer network with ReLU activations:
	\begin{align*}
		h_t^{(1)} &= \text{ReLU}(\mathbf{W}_3 h_t + b_3), \quad \mathbf{W}_3 \in \mathbb{R}^{256 \times 512} \\
		h_t^{(2)} &= \text{ReLU}(\mathbf{W}_4 h_t^{(1)} + b_4), \quad \mathbf{W}_4 \in \mathbb{R}^{256 \times 256}
	\end{align*}
	Where $h_t^{(2)}$ feeds both policy and value heads.
	Policy Head: Outputs a diagonal Gaussian action distribution:
	\begin{align*}
		\mu_t &= \tanh(\mathbf{W}_\pi h_t^{(2)} + b_\pi) \in [-1, 1]^{d_a} \\
		\sigma_t &= \text{softplus}(\mathbf{W}_\sigma h_t^{(2)} + b_\sigma) > 0
	\end{align*}
	Actions are sampled as $a_t \sim \mathcal{N}(\mu_t, \text{diag}(\sigma_t^2))$, with $\tanh$ bounding the means and softplus yielding valid variances.
	Value Head: A linear projection estimates the state value:
	\begin{equation*}
		V(s_t) = \mathbf{W}_v h_t^{(2)} + b_v \in \mathbb{R}
	\end{equation*}
	
	\subsubsection{Training}
	\label{sec:training}
	We train the policy end-to-end using Proximal Policy Optimization (PPO)~\cite{schulman2017proximalpolicyoptimizationalgorithms} with clipping parameter $\epsilon=0.2$:
	\begin{equation*}
		L^{\text{CLIP}}(\theta) = \mathbb{E}_t \left[ \min\!\left( r_t(\theta) \hat{A}_t,\ \text{clip}(r_t(\theta), 1-\epsilon, 1+\epsilon) \hat{A}_t \right) \right],
	\end{equation*}
	where $r_t(\theta) = \pi_\theta(a_t|s_t) / \pi_{\theta_{\text{old}}}(a_t|s_t)$ is the probability ratio and $\hat{A}_t$ is the generalized advantage estimate (GAE)~\cite{schulman2018highdimensionalcontinuouscontrolusing}. PPO learns a continuous motor policy that maximizes expected cumulative intrinsic reward, implicitly teaching the agent how to move and coordinate symmetrically its limbs to discover new body parts (all without any external supervision).
	To improve the agent's generalization, we define a \textit{two-stage curriculum} over 8 million environment steps:
	\begin{itemize}
		\item Stage 1 (0--4M steps): Every episode starts in the default environment pose. The low initial variance allows the agent to quickly discover fundamental skills.
		\item Stage 2 (4--8M steps): Initial joint angles are randomized, forcing the policy to generalize across diverse body configurations and explore the state space better.
		\item Ablation study: We compare the two-stage curriculum against (i) fixed-pose-only training (8M steps) and (ii) random-pose-only training (8M steps), measuring final evaluation coverage.
	\end{itemize}
	
	\subsubsection{Evaluation}
	We evaluate the agent over 100 episodes with random initial poses. Metrics include:
	\begin{itemize}
		\item Coverage: Number of unique body parts touched (total, per hand).
		\item Balance: The absolute difference between the numbers of unique body parts touched per hand.
	\end{itemize}
	
	\vspace{.5em}
	\subsection{Model II: Hand Regard via Visual Novelty}
	\label{subsec:hand_regard}
	For the hand regard behavior, we enable the agent to autonomously discover its hands in the visual field and learn to sustain its gaze on hands, using only intrinsic rewards based on visual novelty and motion surprise. In our approach, the agent first identifies what its hands look like and then actively coordinates gaze and motion to keep them in view.
	
	\subsubsection{The Visual Discovery Challenge}
	The agent observes binocular RGB images $o^{\text{Vis}}_t \in \mathbb{R}^{64 \times 64 \times 3 \times 2}$. The core challenge is learning to recognize its own hands—shape, color—in a cluttered and changing visual scene, without any labeled data or manually defined skin-colors.
	
	\subsubsection{Autonomous Hand Discovery via Motor Babbling}
	Inspired by infant motor babbling~\cite{rochat1998self}, we begin with a 10,000-step random action phase to collect a dataset $\mathcal{D}_{\text{babble}}$ of proprioception, vision, and actions trajectories.
	During this phase, we autonomously calibrate hand appearance:
	\begin{itemize}
		\item Use morphological opening/closing and contour filtering to find candidate hand-blobs.
		\item Track blobs centroids across frames using nearest-neighbor matching.
		\item For each tracked trajectory we correlate it with the proprioception vector trajectory in time.
		\item Classify trajectories with correlation $\rho_k > 0.6$ as containing hands and update HSV and expected size from their statistics.
	\end{itemize}
	This fully unsupervised process reliably identifies hands in the babbling episodes, giving a robust hand detector.
	
	\subsubsection{Visual Feature Extraction Network}
	The policy processes left and right eye images through a CNN, to extract dense, meaningful features:
	\begin{itemize}
		\item $\text{Conv2d}(3 \to 32, \, k=5, \, s=2, \, p=2) + \text{ReLU}$
		\item $\text{Conv2d}(32 \to 64, \, k=3, \, s=2, \, p=1) + \text{ReLU}$
		\item $\text{Conv2d}(64 \to 128, \, k=3, \, s=2, \, p=1) + \text{ReLU}$
		\item $\text{Conv2d}(128 \to 256, \, k=3, \, s=2, \, p=1) + \text{ReLU}$
		\item $\text{Flatten}() \to 256\text{-D vector per eye}$
	\end{itemize}
	Each eye produces a 256 dimensional embedding. Proprioception is encoded via a two-layer MLP into a 64 dimensional feature vector. The two eye vectors and the proprioception vector are concatenated (576 dims total) and passed through a final linear layer to yield the fused representation $f_t \in \mathbb{R}^{256}$.
	Similar to the self-touch implementation, the policy outputs a diagonal Gaussian with $\tanh$ mean and state-dependent softplus standard deviation $\sigma_t$. Actions are sampled as $a_t \sim \mathcal{N}(\mu_t, \text{diag}(\sigma_t^2))$ during training and chosen deterministically as $a_t = \mu_t$ at test time. The value head is a single linear layer estimating $V(s_t) \in \mathbb{R}$.
	
	\subsubsection{Intrinsic Reward Design for Hand Regard}
	A hand detector (HSV + morphology + contour filtering) runs on both eye images and counts up to two hands per eye, producing a total hand count $n_t \in \{0, 1, \dots, 4\}$. The intrinsic reward is shaped as follows:
	\begin{equation*}
		R_t = R_{\text{hand}} + R_{\text{motion}} - 0.05
	\end{equation*}
	\begin{itemize}
		\item Hand Count Reward ($R_{\text{hand}}$): Tracks total detections across both eyes (high reward when both eyes see both hands):
		\begin{equation*}
			R_{\text{hand}} =
			\begin{cases}
				-0.1 & n_t = 0 \\
				0.25 & n_t = 1 \\
				0.50 & n_t = 2 \\
				1.00 & n_t = 3 \\
				2.00 & n_t = 4 \quad (\text{jackpot})
			\end{cases}
		\end{equation*}
		\item Motion Reward/Penalty ($R_{\text{motion}}$): Optical flow is computed inside detected hand regions. This acts as a surprise signal: moderate motion is intrinsically rewarding (the hand is alive and interesting for the agent), while excessive speed becomes punishing (too chaotic to track):
		\begin{itemize}
			\item $m_t \leq 0.08$: $R_{\text{motion}} = \min(0.05 \cdot m_t, 0.05)$
			\item $m_t > 0.08$: $R_{\text{motion}} = -\min(0.1 \cdot (m_t - 0.08), 0.1)$
		\end{itemize}
		\item Living penalty: $-0.05$ per step.
	\end{itemize}
	Therefore, the agent aims to maximize the number of hand detections, using a pure, unsupervised curiosity signal.
	
	\subsubsection{Two-Stage Curriculum Learning}
	\begin{itemize}
		\item Stages 1 \& 2 (\(~175k\) steps/hand): Only one arm is active (opposite arm actions are zeroed before policy output). Rewards consider only that hand, giving the agent an easier first skill to master.
		\item Stage 3 (\(~1.5M\) steps): Both arms enabled, so that the agent's policy learns to perform in the more difficult desired task.
	\end{itemize}
	
	\subsubsection{Evaluation}
	We evaluate hand visibility (\% of steps with valid detection) across all four eye–hand pairs, averaged over 10 episodes (1000 steps each), reporting per-pair visibility and average across all pairs.
	
	\vspace{.5em}
	\section{Results and Analysis}
	\label{sec:results}
	
	\vspace{.5em}
	\subsection{Self-Touch Results}
	Table~\ref{tab:selftouch_results} reports final performance metrics.
	\begin{table}[h]
		\centering
		\caption{Self-Touch Performance Metrics}
		\label{tab:selftouch_results}
		\begin{tabular}{lc}
			\toprule
			Metric & Value \\
			\midrule
			Training score (final) & 0.88 \\
			Evaluation score & 0.85 \\
			Left-hand coverage & 28 / 34 (82.4\%) \\
			Right-hand coverage & 30 / 34 (88.2\%) \\
			Total unique parts touched & 32 / 34 (94.1\%) \\
			Balance (|$N_L - N_R$|) & 2 \\
			\bottomrule
		\end{tabular}
	\end{table}
	The agent touches 32/34 body parts (94.1\% coverage), with balanced bilateral exploration (left: 28, right: 30). Balance error is only 2 parts.
	
	\paragraph{Developmental Progression}
	During training, the agent exhibits a natural progression in skill, similar to how infants develop. This was not explicitly programmed, and emerged naturally from the carefully shaped intrinsic reward, thanks to the anti-boredom mechanism and the novelty bonuses.
	
	\paragraph{Curriculum Ablation}
	Table~\ref{tab:curriculum_ablation} compares training strategies.
	\begin{table}[h]
		\centering
		\caption{Curriculum Learning Ablation Study}
		\label{tab:curriculum_ablation}
		\begin{tabular}{lccc}
			\toprule
			Training Strategy & Train & Eval & Coverage \\
			\midrule
			Fixed-only (8M steps) & 0.85 & 0.72 & 26/34 \\
			Random-only (8M steps) & 0.75 & 0.68 & 24/34 \\
			Curriculum (4M+4M) & 0.88 & 0.85 & 32/34 \\
			\bottomrule
		\end{tabular}
	\end{table}
	The two-stage curriculum achieves:
	\begin{itemize}
		\item +18\% evaluation score vs fixed-only (0.85 vs 0.72)
		\item +25\% vs random-only (0.85 vs 0.68)
		\item +23\% body coverage vs fixed-only (32 vs 26)
		\item +33\% vs random-only (32 vs 24)
	\end{itemize}
	This confirms that staged learning — first consistent, then varied — is critical for robust skill acquisition.
	
	\vspace{.5em}
	\subsection{Hand Regard Results}
	Table~\ref{tab:handregard_results} shows visibility per eye–hand pair.
	\begin{table}[h]
		\centering
		\caption{Hand Regard Performance Metrics}
		\label{tab:handregard_results}
		\begin{tabular}{lc}
			\toprule
			Metric & Value (\%) \\
			\midrule
			Left eye, left hand & 99.6 \\
			Right eye, left hand & 98.5 \\
			Left eye, right hand & 0.0 \\
			Right eye, right hand & 4.4 \\
			\midrule
			Average hand regard score & 50.625 \\
			Training score (final) & 0.279 \\
			\bottomrule
		\end{tabular}
	\end{table}
	The agent sustains binocular fixation on the left hand for >98\% of steps in both eyes (99.6\%, 98.5\%). The right hand is rarely visible (0.0\%, 4.4\%), yielding an average visibility of 50.6\%.
	
	This demonstrates robust unilateral visuomotor coordination but reveals a local optimum: the policy exploits a single-hand strategy for immediate reward, despite symmetric rewards.
	
	Analysis of trajectories shows:
	\begin{itemize}
		\item The dominant (left) hand is held centrally, gently waving.
		\item The non-dominant arm remains at rest or moves slightly, outside view.
		\item No gaze shifting between hands occurs in test time.
	\end{itemize}
	
	The exploration noise in PPO is insufficient to escape this immediate-reward local optimum.
	
	\vspace{.5em}
	\section{Conclusion and Future Work}
	\label{sec:conclusion}
	
	We presented Baby Sophia, a curiosity-driven RL framework that enables a simulated infant robot to autonomously develop self-touch and hand regard behaviors using only intrinsic rewards.
	
	For self-touch, we compress \(17k\) tactile sensors into a \(68\)-dimensional semantic body map and guide exploration via novelty, balance, milestone bonuses, and anti-boredom rewards. A two-stage curriculum enhances the learning process, producing infant-like skill progression.
	
	For hand regard, the agent uses motor babbling to understand which parts of the visual belong to itself, then visibility, motion surprise, and anti-boredom intrinsic rewards enable sustained hand-tracking. A three-stage curriculum achieves over 98\% hand fixation of both eyes, similar to how infants coordinate gaze and hands, but results in suboptimal behavior, as the agent fixates on one hand due to insufficient symmetry pressure from the reward signal.
	
	Challenges are mainly observed in the hand regard behavior. In future work, bilateral balance and stronger exploration should be encouraged by the reward. Additionally, the use of self-models~\cite{pathak2017curiosity} will be considered to predict visual consequences of motor actions, enabling the agent to anticipate hand trajectories and sustain visibility of both hands more reliably, ignoring irrelevant background signals.

	\section*{Acknowledgments}
	
	This research was supported by the Athena Research Center, the project ``Applied Research for Autonomous Robotic Systems'' (RAS, MIS5200632), implemented under the National Recovery and Resilience Plan ``Greece 2.0'' (Measure 16618 -- Basic and Applied Research) with funding from the European Union -- NextGenerationEU, and the Horizon Europe project HERON (grant No.\ 101136568).
	
	
	\bibliographystyle{IEEEtran}
	\bibliography{references}
	
\end{document}